\documentclass[10pt,conference,final]{IEEEtran}
\def\IEEEtitletopspace{18pt}
\IEEEoverridecommandlockouts

\IEEEoverridecommandlockouts

\usepackage{algorithm}
\usepackage[noend]{algorithmic}
\usepackage{amsmath}
\usepackage{amssymb}
\usepackage{color}
\usepackage{xcolor}
\usepackage{graphicx}
\usepackage{dblfloatfix}
\usepackage{multicol}
\usepackage{multirow}
\usepackage{subcaption}
\usepackage{tabularx}
\usepackage{times}
\usepackage{siunitx}

\usepackage{enumitem}
\usepackage{xurl}
\usepackage{adjustbox}
\usepackage{titlesec}

\makeatletter
\let\NAT@parse\undefined
\makeatother
\usepackage{hyperref}
\hypersetup{
    colorlinks=true,
    linkcolor=orange,
    filecolor=magenta,
    urlcolor=orange,
    citecolor=orange,
}
\usepackage{pdfx}

\usepackage[numbers]{natbib}

\usepackage{subfiles}

\renewcommand{\baselinestretch}{0.98}
\titlespacing\section{0pt}{6pt plus 2pt minus 2pt}{0pt plus 2pt minus 2pt}

\begin{document}

%
\title{In-Mouth Robotic Bite Transfer with\\ Visual and Haptic Sensing}

\author{Lorenzo Shaikewitz$^{1,*}$, Yilin Wu$^{2,*}$, Suneel Belkhale$^{2,*}$, Jennifer Grannen$^{2}$, Priya Sundaresan$^{2}$, Dorsa Sadigh$^{2}$\\ \text{\footnotesize California Institute of Technology$^{1},\,
$Stanford University$^{2}$. $^*$ denotes equal contribution.}}

\maketitle

\begin{abstract}
Assistance during eating is essential for those with severe mobility issues or eating risks. However, dependence on traditional human caregivers is linked to malnutrition, weight loss, and low self-esteem. For those who require eating assistance, a semi-autonomous robotic platform can provide independence and a healthier lifestyle. We demonstrate an essential capability of this platform: safe, comfortable, and effective transfer of a bite-sized food item from a utensil directly to the inside of a person’s mouth. Our system uses a force-reactive controller to safely accommodate the user’s motions throughout the transfer, allowing full reactivity until bite detection then reducing reactivity in the direction of exit. Additionally, we introduce a novel dexterous wrist-like end effector capable of small, unimposing movements to reduce user discomfort. We conduct a user study with 11 participants covering 8 diverse food categories to evaluate our system end-to-end, and we find that users strongly prefer our method to a wide range of baselines. Appendices and videos are available at our website: \url{https://tinyurl.com/btICRA}.
\end{abstract}

\begin{IEEEkeywords}
Physical Human Robot Interaction, Physically Assistive Devices, Human Robot Collaboration
\end{IEEEkeywords}


\section{Introduction}
\label{sec:intro}

Robots have the potential to assist people in many aspects of their daily lives. Eating is an integral part of our day, and yet millions of people, from young children to hospitalized patients and the elderly, require assistance from caregivers to perform this essential function. This includes nearly 70\% of hospitalized elderly and at least a quarter of nursing home patients \cite{Shune2018}. However, many of those who benefit from assistance suffer from a loss of independence that can reduce their sense of self-worth, and may lead to malnutrition and weight loss \cite{Shune2018}. For these individuals, robotic solutions that automate feeding promise a return to independence and a healthier lifestyle.

\begin{figure} 
\centering
\includegraphics[width=\linewidth]{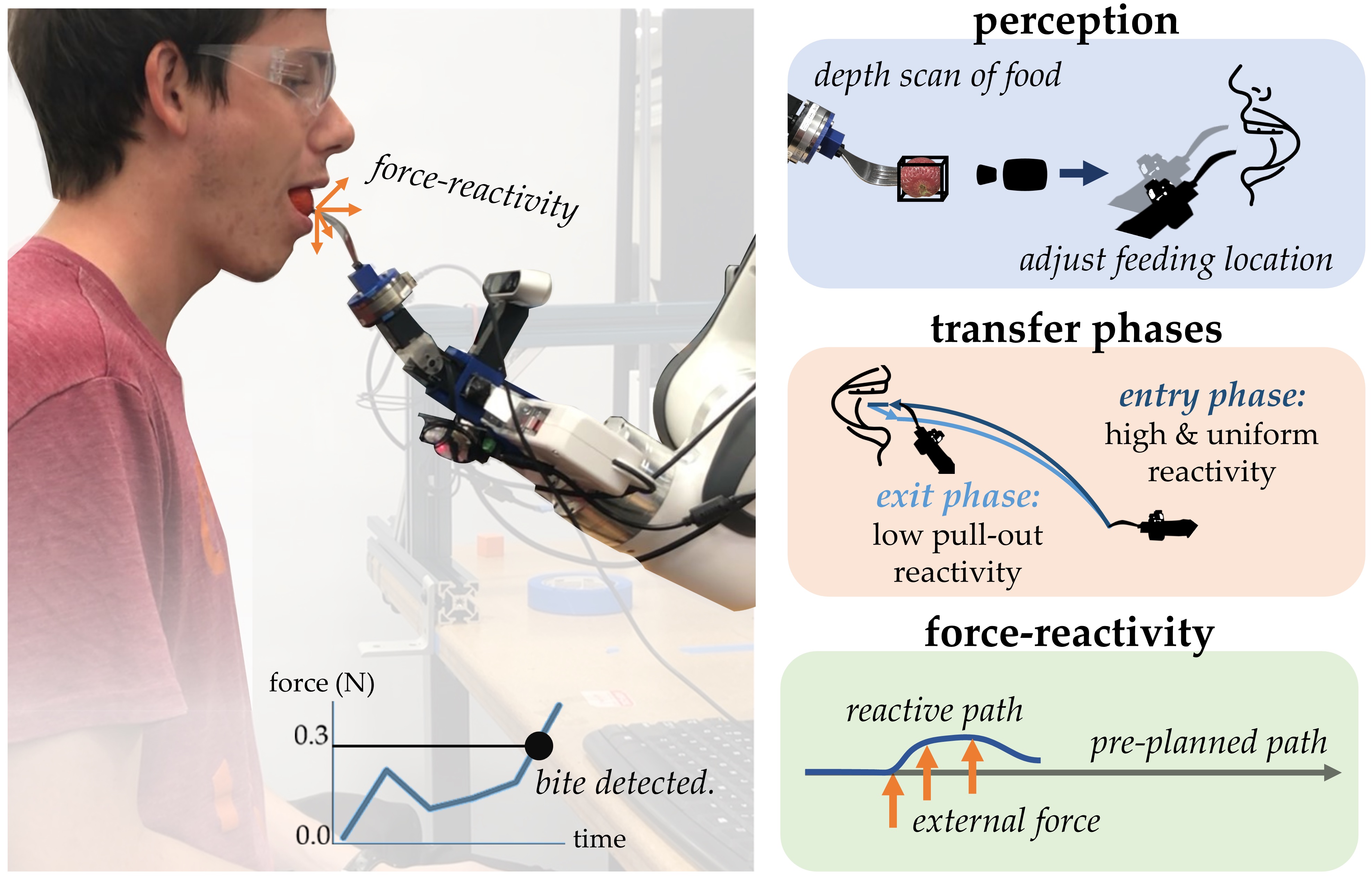}
\caption{\textbf{Robust Robotic Bite Transfer.} Our system safely and comfortably feeds a user bite-sized food items. For an acquired food item, the system detects the user's mouth and adjusts its target position based on a depth scan of the food. To feed the user, it follows an arced trajectory and monitors force to detect a bite. Throughout feeding, it employs a force-reactive controller for user comfort and safety.}
\label{fig:front}
\vspace{-0.5cm}
\end{figure}

One critical piece of the feeding process is robot-assisted \textit{bite transfer}: bringing various food items into a user's mouth using a tool, allowing them to safely take a bite, and then removing the tool from the mouth (see Figure \ref{fig:front}). While several works have studied fork-based bite transfer, most works bring the food item to a point outside the user's mouth and then have the user move forward to bite the food on their own \cite{park2016feedinggeneralmanipulator}. This is simply not practical for the 300,000 people in the U.S. alone who suffer from spinal cord injuries and \textit{require} caregivers to feed them, nor is it comfortable even for the broader population of people with motor impairment disabilities \cite{tapo2020moreautonomy}. The \textit{in-mouth} phase of bite transfer is less studied, and it is particularly challenging since we are bringing a tool (often a fork) inside one of the most fragile and sensitive parts of the human body. It is also difficult to model the human mouth and its actuation, nor do we know when or how the human wants to take a bite. Furthermore, the robot needs to be able to handle a wide diversity of foods and food orientations on the tool. Most recently, \citeauthor{Belkhale2021} developed an in-mouth transfer system which balances comfort and bite efficiency, where comfort was defined as encroaching on a user's personal space when approaching the mouth. They utilize a single reactive controller for adapting the end effector pose to the forces the user applies on the utensil. While this method improves comfort for transfer when \textit{approaching} the mouth, this work does not address user comfort during the in-mouth transfer process. 


There are several key ways in which prior work fails to address comfort during in-mouth transfer. Firstly, prior work uses a single, unchanging reactive controller for the entire bite transfer process. It is important to be reactive to the person being fed; however, the type of reactivity changes based on the \textit{phase} of bite transfer. For example, when feeding a strawberry to a user, we might want to be very compliant during the entry into the mouth, stiff during the bite, and then stiff in the direction of exit in case the user wants to pull the strawberry off the fork. A single reactive controller would fail to take this dimension of comfort into account. Secondly, prior works directly mount tools to the robot end effector for transfer, which can force the robot to make large joint position movements just to adjust the tool orientation. Often, this motion at robot joints early in the kinematic chain leads to large movements during transfer, which is jarring and uncomfortable for users \cite{Belkhale2021}. This is analagous to a caregiver reorienting their elbow, shoulder, and even torso while they are feeding someone, as opposed to subtle movements at the wrist. Thirdly, the prior in-mouth bite transfer work evaluates their method only on carrots \cite{Belkhale2021}. It is necessary to test on more diverse and complex food items (e.g., deformable foods like cheesecake, irregularly shaped foods like broccoli) to draw meaningful conclusions about comfort for in-mouth transfer.

In this work, we develop a comfortable 
system for in-mouth bite transfer across a diverse set of bite-sized foods. To enable comfortable in-mouth transfer of diverse food items, this work builds on two key insights. First, we posit that bite transfer can be split into two phases: mouth entry and mouth exit, separated by bite detection. We can then design \textit{phase-specific} reactive controllers enabling comfortable physical interactions with the robot during the in-mouth transfer. During mouth entry, our system obtains precise measurements of the food item and the user's facial keypoints using vision, and then utilizes this information to bring the food item into the user's mouth. During entry and exit, the system uses a reactive controller to avoid sensitive collisions with the user's mouth, where the specific reactivity coefficients can be different for each phase. Second, inspired by the mobility of the human wrist, we aim to enable more degrees of freedom at the end effector to reduce the jarring motion caused by robot joints earlier in the kinematic chain. To accomplish this, we develop a custom ``wrist" joint to enable fine-grained reactivity while in the mouth.
Using this full system, we conduct a user study with 11 participants and 8 different bite-sized food items to evaluate our system. Our method is substantially preferred over the baselines, and we perform an in-depth analysis of qualitative and quantitative user feedback to guide future work.

\section{Related Work}


\subsection{Robot-Assisted Feeding}
Robot-assisted feeding has become an active area of research and development in recent years. There are several robotic feeding systems on the market currently; however, they lack widespread adoption due to minimal autonomy or dependence on food preparation beforehand \cite{obi, myspoon, mealmate, mealbuddy, gemici2014learning}. In the literature, the autonomous feeding problem is generally split into two phases: bite acquisition and bite transfer.

\smallskip \noindent\textbf{Bite Acquisition}: Within bite acquisition, \citeauthor{2019Feng} developed the SPANet network, which uses supervised learning to map segmented foods of known classes to a corresponding set of discrete actions \cite{2019Feng}. \citeauthor{gordoncontextual} added an online contextual bandits framework for updating the SPANet action prediction using haptic information \cite{gordoncontextual}. \citeauthor{sundaresan2022learning} introduce a probing action during acquisition to zero-shot predict the optimal acquisition action given both visual and haptic information, without the need for multiple trials, across a wide range of foods \cite{sundaresan2022learning}. Beyond fork-based acquisition, \citeauthor{grannen2022learning} develop a spoon-based acquisition framework that leverages two robot arms to provide stability during scooping \cite{grannen2022learning}. However, none of these approaches consider the problem of bite transfer once the food is acquired. Recently, \citeauthor{2019Gallenberger} demonstrate that bite transfer indeed depends on bite acquisition, but they do not provide an approach for feeding bites of food in a general fashion \cite{2019Gallenberger}.

\smallskip \noindent\textbf{Bite Transfer}: Similar to acquisition, autonomous systems for bite transfer usually rely on discrete food classes with predetermined feeding trajectories \cite{2019Feng, 2019Gallenberger, 2016Park, herlant_thesis}. These works also generally bypass in-mouth transfer by navigating the utensil to a predetermined pose outside of the user's mouth, requiring the user to reach the utensil on their own to complete the feeding process. One recent work demonstrated an in-mouth transfer system that optimized for comfort and efficiency during planning; however, this work primarily studied comfort during the \textit{approach} trajectory and did not consider the in-mouth comfort \cite{Belkhale2021}. These works also utilize rigidly mounted utensils for feeding, leading to jarring motion of the robot during transfer as described in Sec.~\ref{sec:intro}. Each of these prior works utilizes a limited set of foods for bite transfer evaluation. Our work aims to specifically improve user comfort during in-mouth transfer.

\subsection{Physical Human-Robot Interaction}
Physical human-robot interaction encompasses scenarios where the human and robot must physically interact in the real world in a safe manner. Prior work studies how robots can safely respond to such interaction \cite{haddadin2016physical,de2008atlas,ikemoto2012physical}, including the use of reactive controllers \cite{haddadin2008collision,okunev2012human}. Several works look at how robots can learn from physical interaction, for example through physical corrections \cite{li2021learning, bajcsy2017learning}. Bite transfer is a specific type of physical human robot interaction. However, bite transfer involves safety and comfort in a constrained and sensitive workspace, a human's mouth, where it is critical for the robot to be precise and reactive to the user. Most closely related to the bite transfer problem is robot-human handovers, which usually involve the passing of some item from a robot arm to a human hand. These works do not usually consider using an intermediate tool for handover~\cite{agah1997humanInteractionRobot, edsinger2007cooperativeManipulation, huber2008handingOverTasks}. Unlike in the feeding domain, handovers usually involve a person's hand as the recipient. \citeauthor{cakmak2011human} find clear human preferences for object orientations and grasp placement in the application-agnostic handover problem. However, passing an item to a person's mouth presents a host of new challenges involving safety and comfort. \citeauthor{gerard2016personalization} show the relevance of individual preferences to the bite transfer process. In our work, we consider the in-mouth handover of bite-sized food items, and we study how to make such transfer both safe and comfortable.

\section{In-Mouth Bite Transfer System}
This section details the operation and design of our robotic bite transfer system. We begin with a description of the hardware platform in Section \ref{sec:eedesign}, describing the general purpose robotic arm and our custom-built actuated wrist-like end effector. Then, in Section \ref{sec:reactive_design}, we present our phased force-reactive controller, which provides a basic measure of safety and enhances comfort by adjusting during the feeding process. Next, Sections \ref{sec:target_design} and \ref{sec:traj_design} detail the mechanisms of the bite transfer system. In Section \ref{sec:eedesign}, we explain how the system chooses a target location using a depth camera image and face detection. The arced trajectory to reach this target and the bite detection system are described in Section \ref{sec:traj_design}. By combining a dexterous end effector and force-reactive control with a robust targeting and transfer procedure, our system safely navigates into the user's mouth and withdraws upon a bite. An overview of this system is shown in Figure~\ref{fig:front}.

%
\subsection{Actuated Wrist-Like End Effector} \label{sec:eedesign}
\begin{figure} 
\vspace{-0.3cm}
\centering
\includegraphics[width=0.95\linewidth]{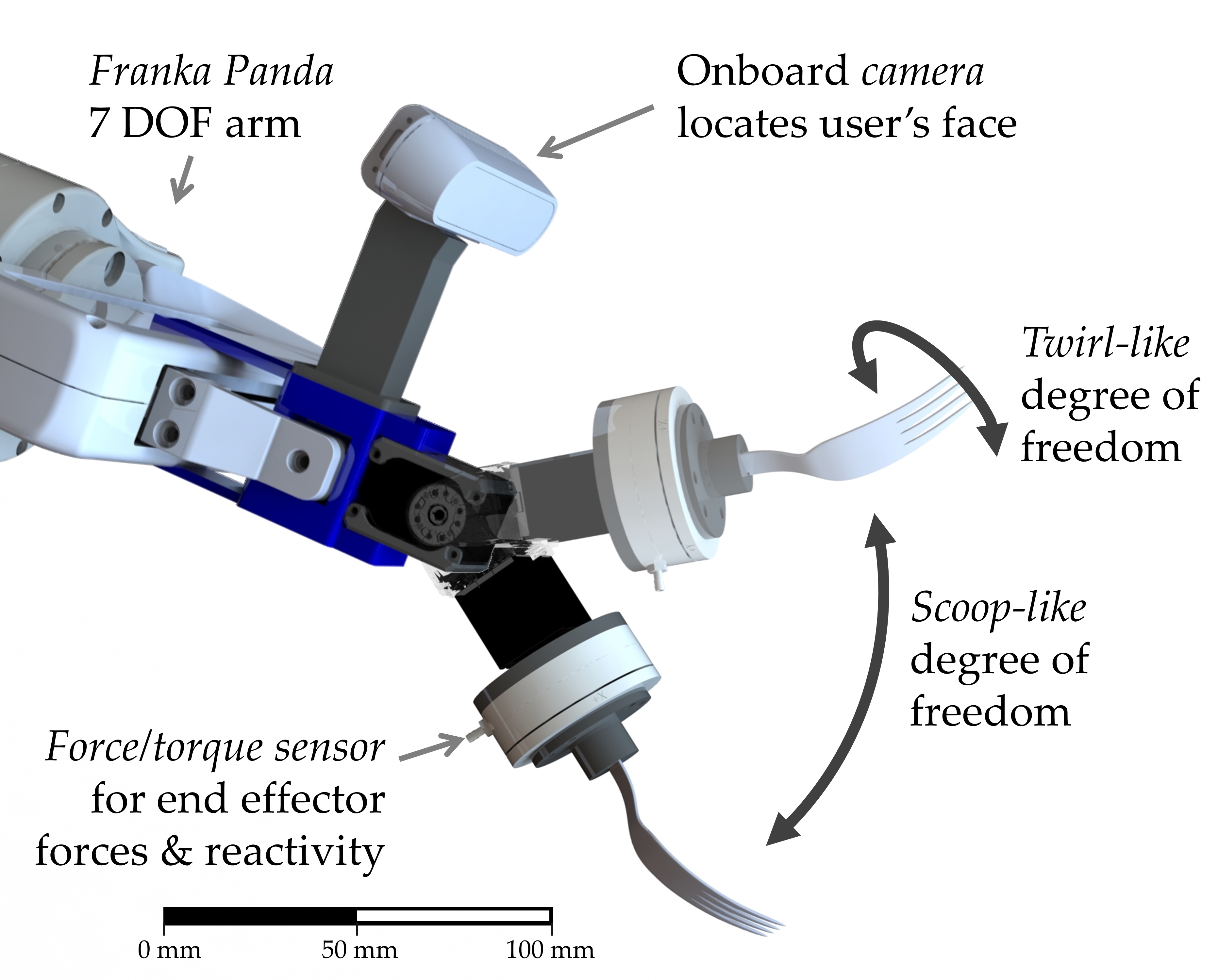}
\caption{\textbf{A Wrist-Like End Effector.} We augment the Franka Panda's gripper with a 2 degree-of-freedom (DOF) end effector designed for feeding motions. This allows simple scooping and twirling motions without moving the entire arm. The wrist also contains perception hardware (mounted RGB-D camera and force/torque sensor) for adjusting to the user's movements.}
\label{fig:wrist}
\vspace{-0.3cm}
\end{figure}

Our system relies on a 7 degree-of-freedom general purpose manipulator (Franka Emika Panda) gripping a custom-built actuated fork assembly. The fork assembly, shown in detail in Figure \ref{fig:wrist}, adds two degrees of freedom designed to mimic the primary movements of a wrist: twirl-like rotation about the fork's long axis, and rotation about a single perpendicular axis to allow a scooping motion. This extends the range of motion of the robot and increases its null space. For feeding, these additions allow rapid adjustment of the fork's position without significant movement of the Panda arm. Fewer movements may also increase feelings of comfort around the arm.

The fork assembly (Figure \ref{fig:wrist}) is composed of two servo motors (Dynamixel XC330-M288-T) connected in series and plastic mounting hardware. It is connected to the Panda's gripper using a 3D printed mount, which also holds a camera (Intel RealSense D435). The twirling degree of freedom is located after the scoop-like degree of freedom at the end of the kinematic chain. A force/torque sensor (ATI Mini45 6-axis F/T sensor) is located between the fork and the second servo motor, forming the tip of the end effector. 

\subsection{Force-Reactive Controller} \label{sec:reactive_design}
Our wrist end effector provides a hardware platform for feeding, but bringing a fork into a user's mouth requires careful consideration of safety. In addition to typical safety features like an e-stop and slow movement, our system is responsive to forces at the tool using a force-reactive controller.

In traditional impedance control, the desired position and velocity is used to make a desired end effector wrench, $f$, which coupled with the Jacobian $J$ gives necessary joint torques $\tau$. Our controller adds reactivity to the forces on the tool, which is equivalent to adjusting the desired wrench before it is passed through the impedance control relation.
Specifically, we add an external force to impedance control: $\tau = J^T \left( f - \bar{f}_m \right)$, where $\bar{f}_m$ comes from a proportional-integral (PI) term on the measured tool forces $f_m$. The PI term is given by $\bar{f}_m = k_P f_m + k_I \int f_m \mathrm{d}t$.




This added reactivity operates at a 1kHz control rate and enables compliance to fine-grained user movements. We tune the sensitivity of this system by adjusting the vector proportionality constants $k_P$ and $k_I$. To prevent unexpected movements from the robot, we set the torque components of these constants to zero. Note that the force reactivity does not extend to the wrist end effector's degrees of freedom, so using torque reactivity requires larger movements.

\smallskip\noindent\textbf{Phased Reactivity.} The sensitivity of the system is key to the safety and comfort of the bite transfer; however, it is not necessary for the reactivity to remain the same in all axes throughout the transfer. Our method incorporates a phased force-reactive controller, using one set of constants as the fork enters the mouth (the \textit{entrance phase}) and a different set of constants as the fork is leaving the mouth (the \textit{exit phase}), as shown in Figure \ref{fig:front}. In the former phase, users benefit from high reactivity because they may safely guide the fork to a comfortable transfer location. Thus, we keep $k_P$ and $k_I$ relatively high ($k_P = 7$ and $k_I = 20 \:\mathrm{Hz}$ for all forces, and $k_P = k_I = 0$ for all torques). During the exit phase, high reactivity can inhibit transfer as the force of a bite prevents successful withdrawal. Therefore, we reduce the reactive constants only in the exit direction ($k_P = 2$ and $k_I = 1 \:\mathrm{Hz}$ for forces into or out of the mouth). This maintains compliance in the directions orthogonal to the exit direction while combating the resistance of the bite for a successful withdrawal. In addition, the reactive constants may be tuned to each individual's preference, which we leave to future work.

\subsection{Defining a Target Location} \label{sec:target_design}
\begin{figure} 
\centering
\includegraphics[width=\linewidth]{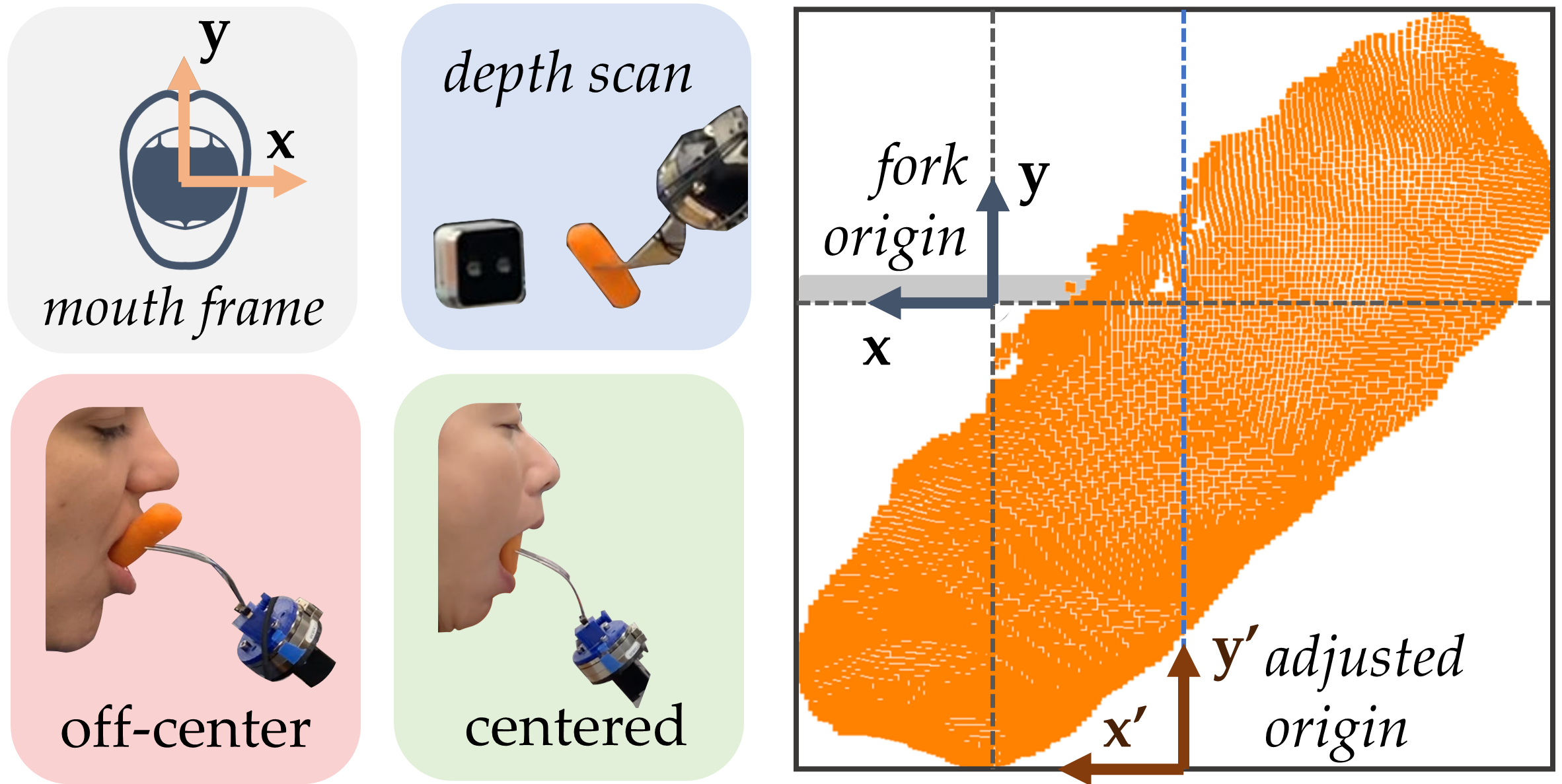}
\caption{\textbf{Adjusting the Target Position.} For any placement of a bite-sized food item on the end of the fork, we compute a safe target position given a depth scan taken from the mouth perspective and centered at optimal entry coordinates for the fork alone. Our system uses the scanned point cloud to recenter and lower the food by averaging the bounding box $x$ coordinates and adjusting by the minimum $y$ coordinate, respectively.}
\label{fig:scanning}
\vspace{-0.3cm}
\end{figure}

As the system is designed to operate with any bite sized food item, we adjust the location where the fork enters the mouth only according to the food's placement on the fork. This is a two phase process: first, an external camera uses a depth scan of the food item to determine its bounding box relative to the fork. Then, a camera mounted on the end effector identifies the location of the user's mouth and applies the offsets computed from the depth scan's bounding box (Figure \ref{fig:scanning}).

\smallskip\noindent\textbf{Offset Generation.} In the depth scan phase, the tines of the fork are facing an external fixed short-range depth camera (Intel RealSense D405). This perspective is ideal for getting offsets in the plane parallel to the user's face; the camera views the end of the fork as if the fork was moving towards it. The resulting point cloud, obtained at $0.1$ mm resolution, is transformed into a ``mouth frame" (Figure \ref{fig:scanning}) located at the tip of the fork. This corresponds to a point just below the teeth in the open mouth: a safe feeding location if there were no food on the fork, and must be adjusted based on food orientation. Specifically, this mouth frame has its $z$-axis pointing out of the mouth and $y$-axis pointing upwards towards the user's eyes. The $x$-axis is along the lip, pointing to the user's left.

From this coordinate system, the axis-aligned bounding box of the point cloud gives the necessary offsets in the face ($x$-$y$) plane. The $x$-axis offset (along the lips) aligns the center of the food, rather than the fork, with the mouth center. This is especially helpful for large, off-centered acquisitions, like the one shown in Figure \ref{fig:scanning}. Assuming the lower jaw does most of the movement to open the mouth, the $y$ direction offset moves the top of the food down to a position where it will not contact the upper teeth. We focus on bite-sized foods in this work so we do not perform a depth measurement to adjust the distance into the mouth that the fork enters. Instead, the distance is fixed to $18$ mm, which we empirically found to be comfortable and safe for most users.

\smallskip\noindent\textbf{Face Detection.} These offsets are added to location of the mouth, which is obtained using a pretrained Histogram of Oriented Gradients (HOG) model that is publicly available through DLib \cite{dlib09}. After the scanning phase is complete, users are instructed to remain still with their mouth closed while the network obtains a mouth center point. This center point is adjusted according to the previously determined offsets to compute the target feeding position. Since the network cannot recognize faces when it gets too close to the user, we rely solely on the initial scan to determine the mouth position; users are expected to remain still throughout the $10$ second trajectory. This assumption, while slightly prohibitive for able-bodied users, is reasonable for many individuals who require feeding assistance but have difficulty moving their heads.

\subsection{Trajectory and Transfer} \label{sec:traj_design}
Throughout the bite transfer process, the trajectory of the robot is designed to maximize comfort by emulating the ways humans feed themselves. After offset and face detection, the end effector moves to a low position and the fork is rotated to an upside down position for easier transfer from below. Then, it follows an arced trajectory (radius $0.45$ m, similar to the length of a lower arm, and centered directly below the target position near the elbow) to a point just in front of the mouth with the fork angled upward. This emulates the motion of an elbow and brings the food in from below, keeping the bulk of the robot arm away from the user. Then, the robot linearly enters the mouth at its target position and slightly lowers the fork to keep it from hitting the top of the user's mouth.

\smallskip\noindent\textbf{Transfer Phases.} Based on the different levels of sensitivity, the bite transfer is divided into two phases, i.e., the entry phase and the exit phase. The system can automatically divide the two phases with a bite detection function. The bite is detected when the magnitude of the force from y axis is greater than the threshold of $0.3$ N (obtained in early bite testing with our system). During the entry phase, the end effector remains motionless and reactive as the system waits for a bite. If no bite is detected within a wait period of $1.5$ seconds, the bite detection times out and begins the exit phase. This relatively short timeout allows the user to reject the food or initiate transfer without a forceful bite. In the exit phase, the fork simply leaves the mouth in a linear trajectory and then follows the same arc path down to its starting position. 

The result is a system capable of robustly feeding a variety of food items directly into a user's mouth. Sampled real-world trajectories are shown in Appendix~\ref{sec:robot_trajectories}\footnote{The appendix may be found at our website: \url{https://tinyurl.com/bticra}}.


\section{Experimental Evaluation}
\label{sec:experiments}
To evaluate the safety, comfort, and overall experience of using our system, we conducted a user study feeding 11 participants food items with our bite transfer system and 4 baselines. We additionally show that our custom wrist end-effector can improve upon several dimensions of user comfort.

\subsection{Experimental Setup}
\noindent
\textbf{Participants and Procedure.}
Our study included 11 able-bodied\footnote{Able-bodied participants were chosen to protect vulnerable individuals amid the ongoing COVID-19 pandemic.} participants and 8 varied food items. Participants (6 female, 5 male, aged between 22 and 57) were somewhat experienced with robots, with 6/11 indicating they had interacted with one before. 7 participants reported they had never been fed as an adult, and 3 reported never having fed someone else. We used a within-subjects design, feeding each subject randomly chosen three food items using each of five methods, twice each. Additionally, we gave each subject one additional feeding trial per method. This study was approved by the Institutional Review Board of Stanford University.

\smallskip\noindent\textbf{Safety Considerations.}
Safety of participants was a primary objective. At each major planning phase (scanning and face detection), the system paused for proctor confirmation before moving towards the user. On top of the compliant reactive controller, we included a software stop: motion was automatically aborted if the end effector force exceeded $3 N$, a mild amount of contact. Additionally, an emergency stop button was in place and available to both the proctor and participant.

\subsection{Study Design}
\noindent\textbf{Independent Variables: Food Settings.}
We evaluate the bite transfer system with 8 food item classes: carrot, strawberry, blueberry, pineapple, cherry tomato, broccoli, cheesecake, and tofu. These food classes cover a wide spectrum of sizes, shapes, and deformabilities (see Table~\ref{table:food_groups} in Appendix~\ref{sec:per_food_rankings}). All food items were cut to bite-size, and we gave participants the opportunity to opt out of eating specific foods.

\smallskip\noindent\textbf{Independent Variables: Feeding Method.}
We compare our method against three robotic feeding baselines and one human oracle baseline. \emph{\textbf{Fixed Pose}} is from prior work and is not an in-mouth bite-transfer method~\cite{2019Gallenberger}. Instead, the robot moves the fork to a position outside the user's mouth and allows the user to take a bite on their own. \emph{\textbf{Less Reactive}} and \emph{\textbf{More Reactive}} are two ablations of our system with less and more force-reactivity respectively. Specifically, the less reactive method uses $k_P = 2$ and $k_I = 2 \:\mathrm{Hz}$ for all forces throughout feeding, while the more reactive method uses $k_P = 10$ and $k_I = 30 \:\mathrm{Hz}$ for all forces. We note that \emph{\textbf{More Reactive}} is representative of the in-mouth bite transfer method in \citeauthor{Belkhale2021}, which uses a single, highly reactive controller. \emph{\textbf{Human Oracle}} involves the user being fed by another human, which serves as a reasonable upper bound for our robotic feeding system.




\smallskip\noindent\textbf{Dependent Measures.} 
We consider both subjective and objective measures.
After each food item was fed with all methods, participants were asked to rank the methods considering safety, comfort, and overall preference (\textit{Safety, Comfort, and Overall Ranking}). After the study, participants provided a \textit{Final Ranking} of the methods based only on their overall preference.
Per feeding method, we measured the fraction of trials in which the bite is successfully transferred (\textit{Success Rate}).


\smallskip\noindent\textbf{Hypothesis.}
We hypothesized that:

\textbf{H1.} By using a multi-phase reactive controller and a more dexterous wrist end-effector, our method will outperform the three robotic baselines in terms of comfort, safety, and overall rankings for all foods.

\textbf{H2.} Our method will have a higher transfer success rate
compared to the three robotic baselines for all foods.

\subsection{Results}


We present both subjective and objective metrics from our user study to evaluate our system.

\begin{figure} [!t]
\vspace{-0.1cm}
\centering
\includegraphics[width=\linewidth]{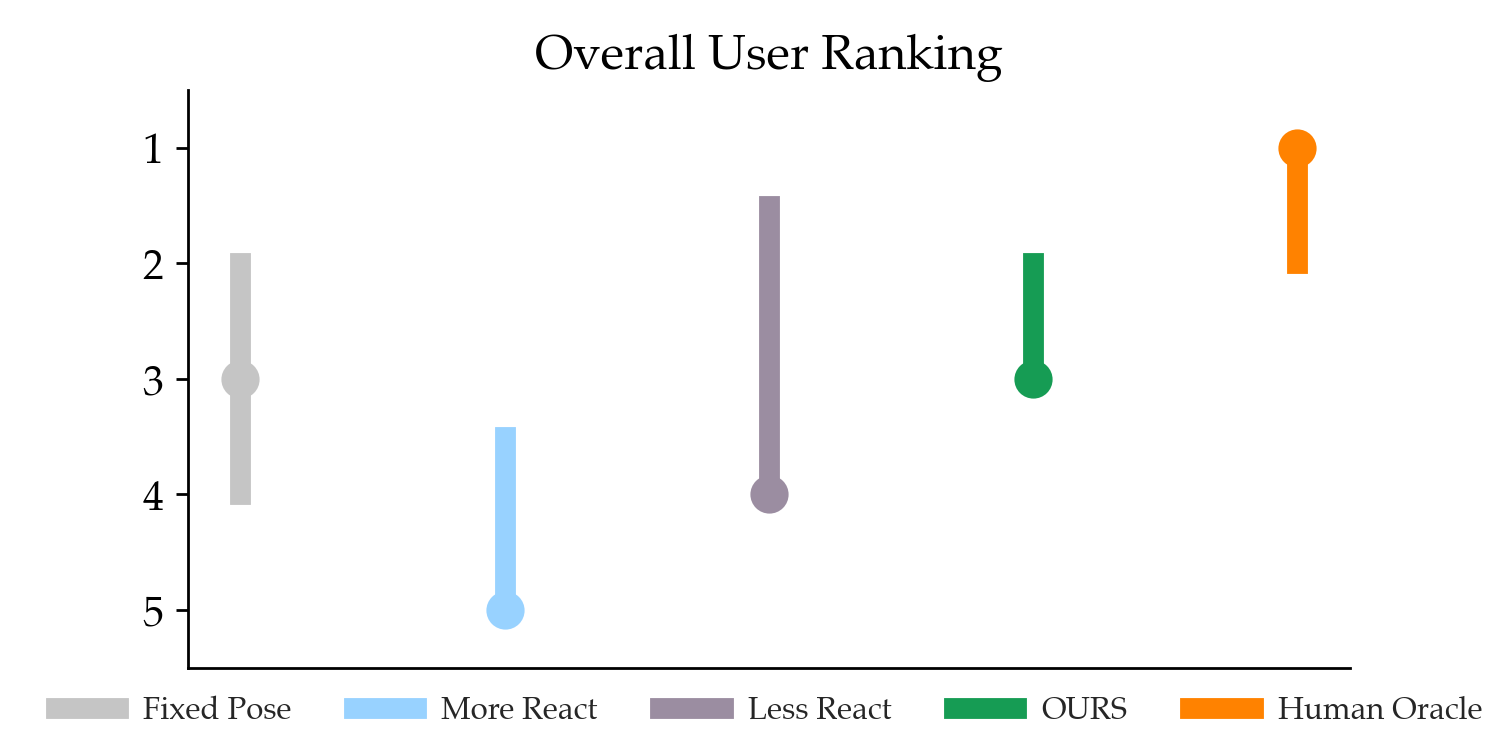}
\caption{\textbf{Overall User Rankings.} We report the median and interquartile range of user rankings across four robotic feeding methods and one human feeding oracle. We compare our method against two similar baselines with varied reactivity coefficients, and one fixed pose baseline that holds the food outside the mouth and places the burden on the user to move to take a bite. Our in-mouth bite transfer has the highest overall user ranking of the robotic feeding methods, suggesting the effectiveness of our phased reactive method.}
\label{fig:overall_rankings}
\end{figure}

\smallskip\noindent\textbf{Subjective Metrics.}
We show the overall user rankings of each of the five feeding methods \textbf{(H1)} in Figure~\ref{fig:overall_rankings}. Additionally, we present user ranking results on the safety, comfort, and overall quality of each feeding method \textbf{(H1)}, grouped by food variability axis (shape, size, and deformability--defined in Appendix~\ref{sec:per_food_rankings}), shown in Figure~\ref{fig:grouped_rankings}. 

\begin{figure}[!h]
\vspace{-0.15cm}
\centering
\includegraphics[width=\linewidth]{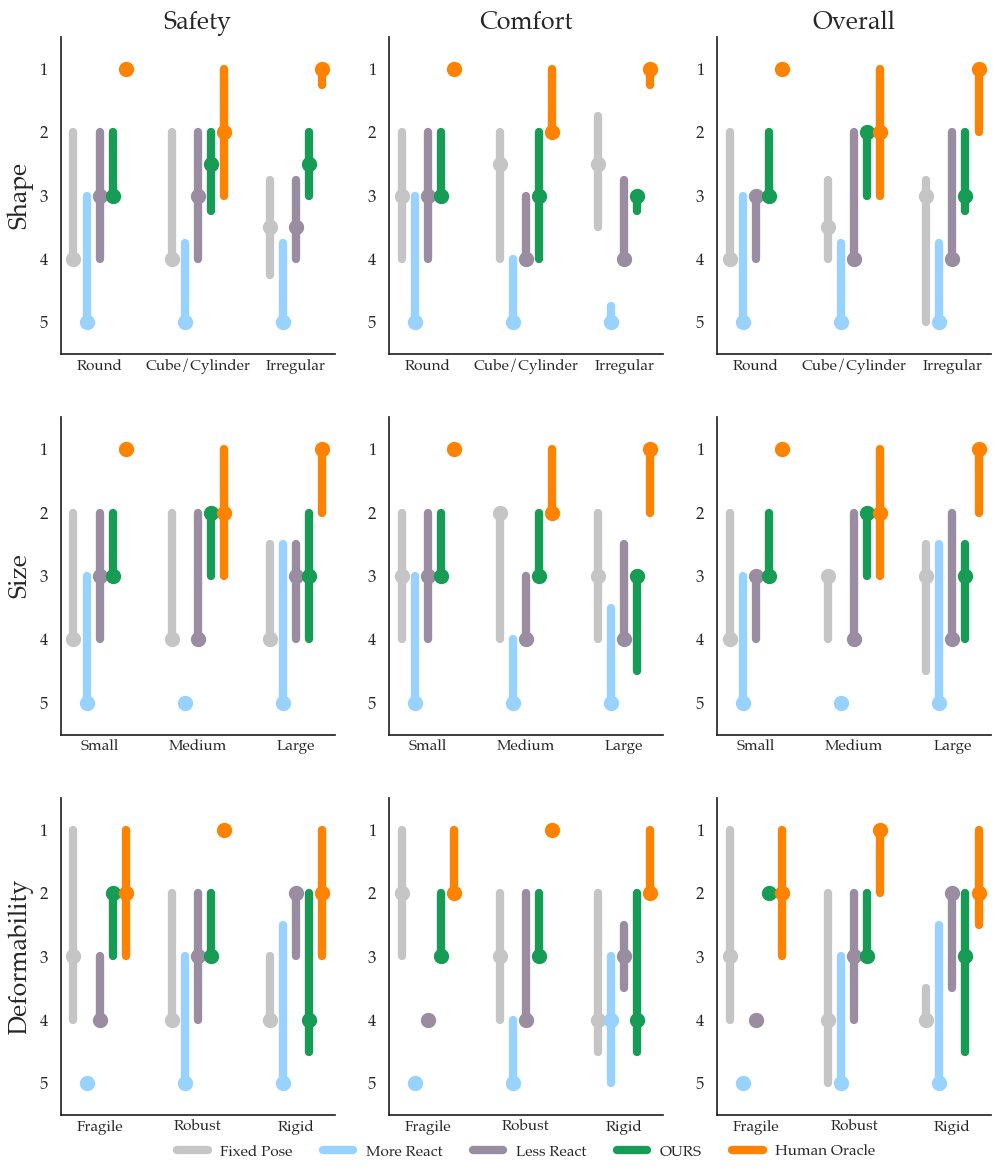}
\caption{\textbf{User Rankings by Food Property.} We report the median and interquartile range of user rankings of the safety, comfort, and overall quality of the feeding method by food property. As shown in Table~\ref{table:food_groups} in Appendix~\ref{sec:per_food_rankings}, we group the 8 foods considered (broccoli, blueberry, cherry tomato, strawberry, carrot, pineapple, tofu, and cheesecake) into 3 categories along each of 3 axes of variability: shape, size, and deformability. Each food item appears exactly once in each variability category. One-way ANOVA testing confirms the statistical significance of transfer style (in-mouth vs. out-of-mouth) and reactivity (less, more, ours) on the reported user ratings for comfort alone, and safety, comfort, and overall, respectively. We report additional per food user rankings and exact ranking values in Appendix~\ref{sec:per_food_rankings}.}
\label{fig:grouped_rankings}
\vspace{-0.3cm}
\end{figure}

Both overall and across food item variabilities, we find that users prefer our in-mouth bite transfer method over all other in-mouth feeding methods (Less Reactive and More Reactive), and that our method performs comparably to a fixed pose transfer method \textbf{(H1)}. This trend holds across safety and comfort ratings as well, as users noted the method ``had a gentle approach and placement of food'' and it was ``easy to bite, the robot reacted and the motion away from the mouth was good''. 
In some instances, our method even approaches a human feeding oracle performance, suggesting that our method consistently performs high-quality bite transfer. 

We find that users rank the fixed pose baseline relatively highly as they are able to ``be active in the feeding process'' and retain more control over their bite, leading to higher ratings for comfort in particular. This is by nature of the method placing the burden on the user to move their head to take a bite off the fork. While this method may be more comfortable for able-bodied users, moving the mouth to the fork may be impractical for those with mobility issues who stand to benefit the most from an assistive feeding robot system. We report additional user rankings on head movement in Appendix~\ref{sec:per_food_rankings}. 

We note that users prefer our in-mouth bite transfer and consider it to be more safe and comfortable than the other robotic feeding baselines across most food groups; in other words, our method of moderate, phased reactivity is generally preferred. One exception is the rigid food class under food deformability, where the Less Reactive baseline is preferred for safety, comfort, and overall. This is due to the high friction of the rigid food against the fork, making it difficult for the user to pull the food off the fork with a more reactive method such as our system or the More Reactive baseline. 

We perform additional one-way ANOVA testing to assess the effects of (1) phased reactivity (\textbf{\emph{Less Reactive}}, \textbf{\emph{Ours}}, or \textbf{\emph{More Reactive}}) and (2) transfer style (in-mouth --- \textbf{\emph{Less Reactive}}, \textbf{\emph{Ours}}, and \textbf{\emph{More Reactive}} --- or out-of-mouth --- \textbf{\emph{Fixed Pose}}) on user ratings. We find that phased reactivity has a statistically significant effect on safety, comfort, and overall ratings, while transfer style has a statistically significant affect on comfort ratings alone (denoted by $p \leq 0.05$), even for a small sample size of $11$ users. The results support the hypothesis (\textbf{H1}) that both phased reactivity and transfer style impact a user's experience. We provide additional details of the ANOVA testing procedure in Appendix \ref{sec:per_food_rankings}.

\begin{table}[h]
\centering
\vspace{-0.2cm}
\begin{tabular}{c | c |c c c}
\multirow{2}{*}{\textbf{Method}} & \multirow{2}{*}{\textbf{Success Rate}} & \multicolumn{3}{c}{\textbf{Failure Mode}}  \\ 
 &  & Bite Failure & Drop & Imprecise \\
\hline
\emph{Fixed Pose} & 65/66 & 0 & 1 & 0 \\
\emph{More Reactive} & 59/66 & 6 & 0 & 1 \\
\emph{Less Reactive} & 63/66 & 2 & 0 & 1 \\
\emph{Ours} & 62/66 & 3 & 1 & 0 \\
\emph{Human Oracle} & \textbf{66/66} & 0 & 0 & 0 \\
\end{tabular}
\makeatletter
\def\@captype{table}
\makeatother
\caption{\textbf{Bite Transfer Success Rate.}
\normalfont{We report the success rate and failure modes of all the methods across 66 trials with 8 different foods and 11 users in the experiments. Three types of failure modes happen in the user study. In \emph{Bite Failure}, users fail to take a bite of the food. This can occur when either the user retracts early due to mistrust or fails to take a bite before the fork exits. In \emph{Drop}, the food falls off the fork during the transfer process. \emph{Imprecise} denotes the face detector localizing the mouth inaccurately, causing a near-miss or off-center transfer. The highest success rate is human oracle and the most common failure mode is Bite Failure.
}}
\label{table:success_rate}
\vspace{0cm}
\end{table}

\smallskip\noindent\textbf{Objective Metrics.}
Our quantitative results measuring bite transfer success \textbf{(H2)} are shown in Table~\ref{table:success_rate}, grouped by failure mode. Among all the methods, \emph{Human Oracle} represents an upper bound on performance. \emph{Fixed Pose} demonstrates the second highest success as it bypasses in-mouth transfer entirely, and consequently any associated challenges. However, this method is a highly impractical solution for people with mobility impairments because it requires the user to significantly move their head to take a bite. Our proposed approach critically alleviates this need with in-mouth transfer, which naturally introduces more potential for imprecision or bite failure. We note that the \emph{More Reactive} method suffers from the most Bite Failure errors, likely caused by jerkiness in the compliant controller which can be jarring to a user. \emph{Ours} and \emph{Less Reactive} perform comparably and outperform the \emph{More Reactive} method, suggesting the effectiveness of phased transfer and less compliance upon fork exit. 
Finally, we report the occurrence of food falling off the fork during the transfer process and mouth detection inaccuracies resulting in imprecise transfer. We note that these minor failures are system-level errors, irrespective of the transfer method employed.



\begin{figure}
    \centering
    \minipage{0.61\linewidth}
        \centering
        \adjincludegraphics[width=\linewidth,trim={.048\width} {0.08\height} {0.05\width} {0.05\height},clip]{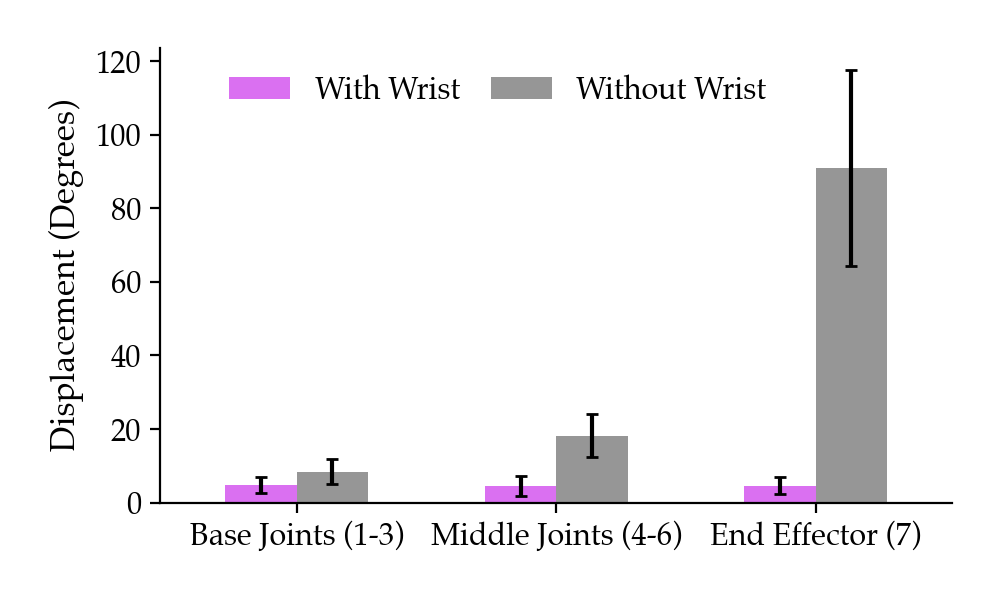}
    \endminipage
    \hfill
    \minipage{0.38\linewidth}
        \centering
        \adjincludegraphics[width=\linewidth,trim={.05\width} {0.08\height} {0.1\width} {0.05\height},clip]{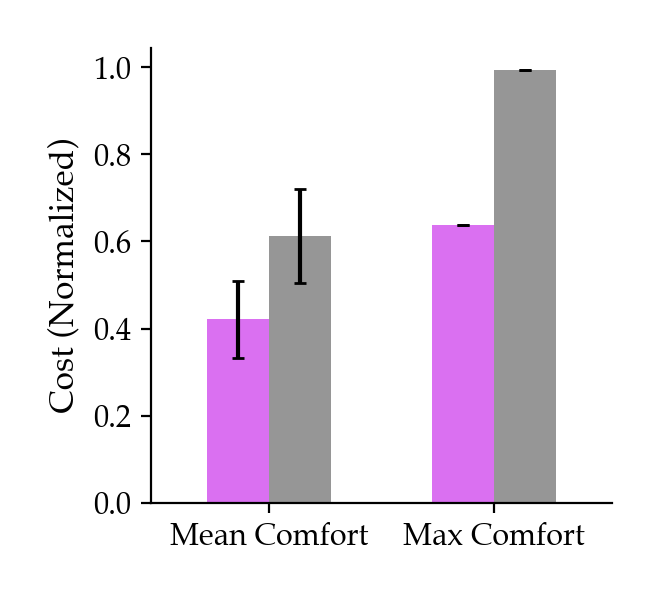}
    \endminipage
    \caption{We generate IK solutions for 10,000 random fork poses using the Wrist joint (With Wrist) as compared to directly mounting the tool (Without Wrist). \textbf{Left}: Average Joint Displacement, for the 7 Franka Panda joints (grouped for simplicity). \textbf{Right}: Average Comfort Cost, defined as encroaching on the user's personal space, similarly to prior work~\cite{Belkhale2021}. Our wrist mount achieves substantially lower joint displacement (less jarring motion) and comfort cost (less personal space occlusion), at no additional computational cost.}
    \label{fig:wrist_displacement_and_comfort}
    \vspace{-0.25cm}
\end{figure}

\smallskip\noindent\textbf{Wrist Joint Analysis.}
In addition to our user study, we analyze and demonstrate the ability of our custom wrist end-effector to improve perceived comfort through simulation experiments in Figure~\ref{fig:wrist_displacement_and_comfort}. We generate 10,000 sampled tool poses that are likely to occur during feeding, and then compute the inverse kinematics (IK) solution for the robot joint configuration using the damped least squares method. We compute IK across two robot setups, one with the tool mounted on our custom wrist end-effector (9 DOF), and one with the tool directly mounted to the robot's default end effector (7 DOF). For both setups, we evaluate the average joint displacement across the 7 robot joints for the sampled tool poses. High joint displacement indicates that the joints earlier in the kinematic chain must move more to reach a given tool pose, thus leading to jarring and uncomfortable motion for the user. By adding more degrees of freedom at the end of the kinematic chain, our wrist-like end effector substantially lowers the average joint displacement (see left of Figure~\ref{fig:wrist_displacement_and_comfort}). Furthermore, we compute a ``comfort" cost similar to the one used in \citeauthor{Belkhale2021} \cite{Belkhale2021}, representing the degree to which the robot encroaches on the user's personal space. By choosing configurations that stay out of a user's field of view, the wrist also lowers the average and maximum personal space comfort cost (see right of Figure~\ref{fig:wrist_displacement_and_comfort}).

\section{Conclusion}

We present a phased bite transfer system that combines visual and haptic information for comfortable and safe in-mouth bite transfer. Our method uses a custom wrist-like joint for additional degrees of freedom and force-reactivity for increased user comfort during feeding. We present results from a user study with 11 participants comparing our in-mouth bite transfer system to baselines of varied reactivity from prior work, fixed pose feeding methods, and a human feeding oracle. We find that our system is rated the highest of the in-mouth robotic feeding methods both in terms of comfort and safety, and is able to significantly reduce robot joint motions that are often jarring to a user using our novel wrist joint.

In future work, we will explore how to incorporate user feedback and preferences in bite transfer to customize a robotic feeding system to each user. Beyond the transfer component, we will consider different trajectories to reduce food droppage during transport, which becomes especially critical for foods that can spill or drip (i.e. soup, salsa).
We will also extend the system to plan for multiple bites of large food items.

\section*{Acknowledgements}
This work was supported by funds from NSF Awards 2132847 and 2218760, as well as the Office of Naval Research.

\bibliographystyle{./bibliography/IEEEtranN}
\bibliography{./bibliography/bib}

\newpage 

\begin{figure*}[!htbp] 
\centering
\includegraphics[width=\linewidth]{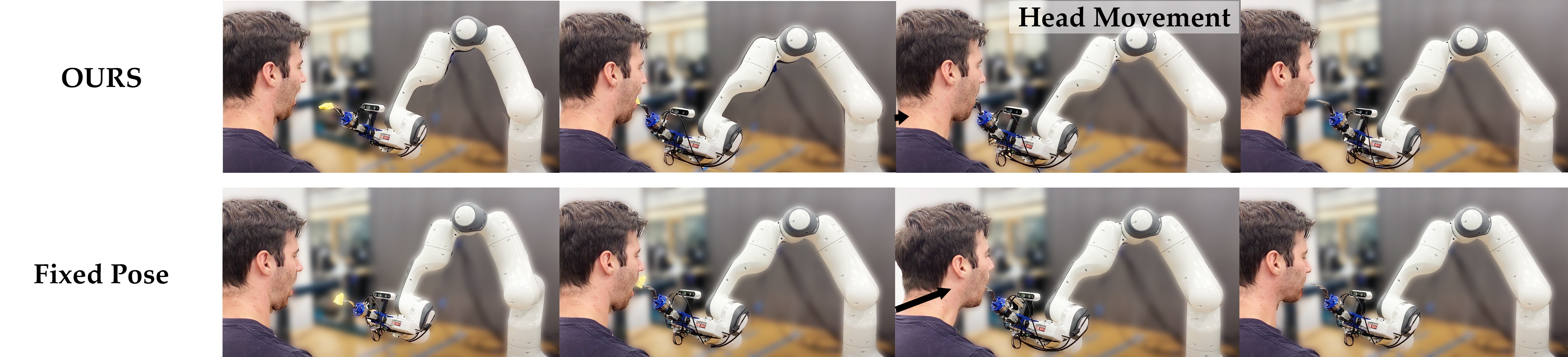}
\caption{\textbf{Sample Trajectories.} We show two transfer trajectories for pineapple with \textbf{\emph{Ours}} and \textbf{\emph{Fixed Pose}} methods. With reactive, \emph{in-mouth} transfer, \textbf{\emph{Ours}} requires less head movement from the user to successfully transfer the bite compared to the \emph{out-of-mouth} \textbf{\emph{Fixed Pose}} strategy. }
\label{fig:sample_trajs}
\end{figure*}

\newpage
\appendix
\maketitle
\label{sec:appendix}
We provide additional details regarding our approach and experimental evaluation in the sections below. Appendix \ref{sec:robot_trajectories} highlights the visual differences in transfer trajectory across methods. Appendix \ref{sec:per_food_rankings} supplements the results from Section \ref{sec:experiments} with a detailed breakdown of user rankings by food type and measured user movement during transfer to further gauge overall, safety, and comfort preferences across methods. Finally, Figure \ref{fig:sample_trajs} includes visualized rollouts across methods, to further highlight the perceptible differences in transfer. Datasets, code, and videos can be found on our \href{https://tinyurl.com/btICRA}{website}.

\subsection{Sample End Effector Trajectories}

\begin{figure}[!htbp]
    \centering
    \includegraphics[width=\linewidth]{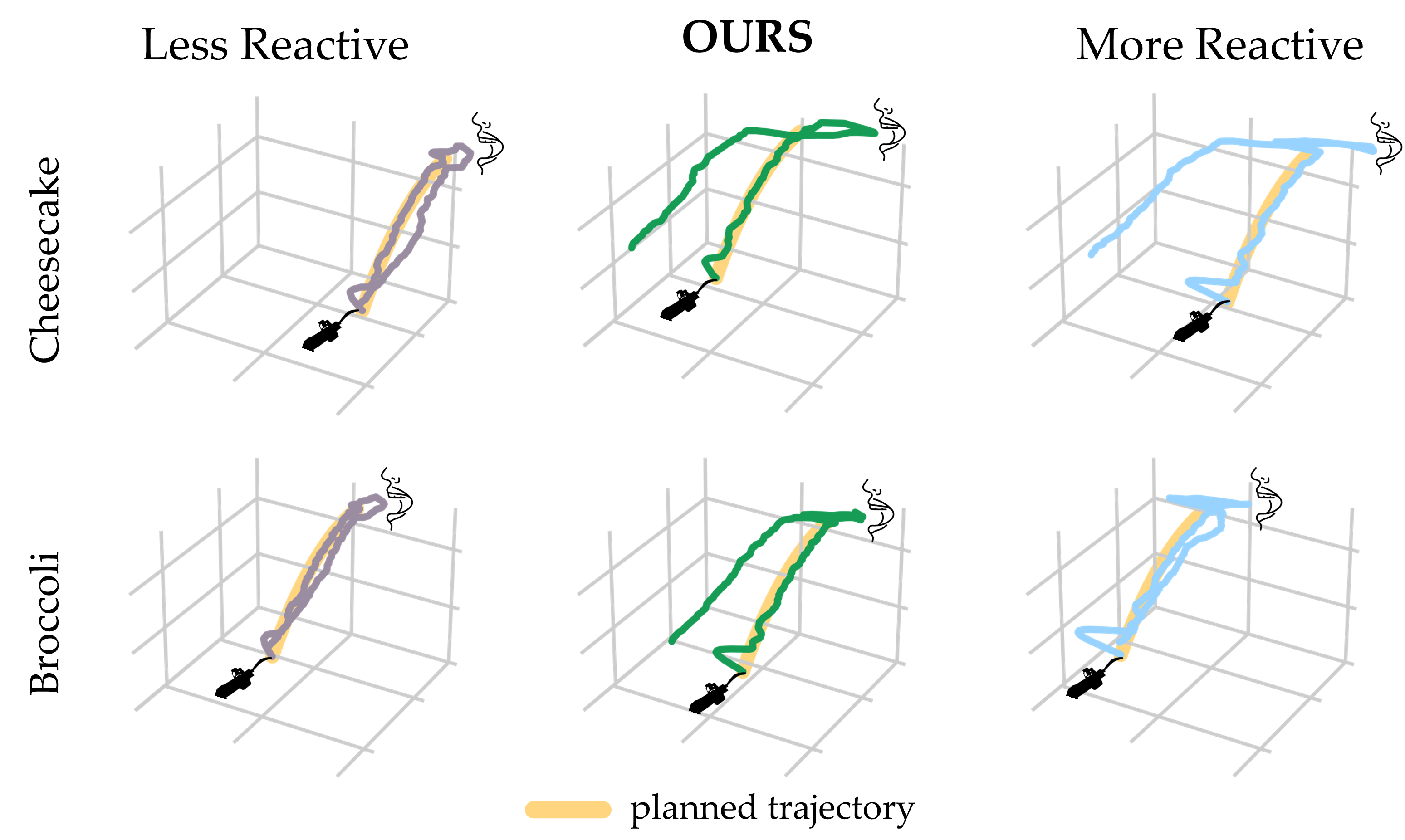}
    \caption{\textbf{End-Effector Trajectories.} Sample end effector position trajectories over the course of a feeding run for the three in-mouth robotic methods. The less reactive trajectory is closest to the pre-planned one, indicating less compliance. In contrast, the more reactive trajectory diverges from the planned one in a jittery path, matching our observations during the trial. Our method balances the two, responding to some user movement without experiencing significant jitter.}
    \label{fig:ee_trajs}
\end{figure}

A difference between the three levels of reactivity evaluated in this paper is clearly seen in the end effector trajectories (see Figure~\ref{fig:ee_trajs}). As expected, the less reactive trajectory sticks to its pre-planned path the closest, while the more reactive one jitters away from it. We show trajectories from individual trials because of different user mouth locations and food items between trials. Note that this result is purely qualitative.

\label{sec:robot_trajectories}

\subsection{Additional User Study Details}
\label{sec:per_food_rankings}

In this section, we report additional results from the user study comparing our bite transfer method to three baselines and a human-feeding oracle. 

We analyzed our user ranking data, in the range of 1-5 across the safety, comfort, and overall categories, using one-way ANOVA testing to measure the statistical significance of both phased reactivity and transfer style. For phased reactivity, we split the into the categories of \emph{\textbf{Less Reactive}}, \emph{\textbf{More Reactive}}, and \emph{\textbf{Ours}}. For transfer style, we group the \emph{\textbf{Less Reactive}}, \textbf{More Reactive}, and \emph{\textbf{Ours}} rankings as the \emph{in-mouth} category, and compare this against the \emph{out-of-mouth} \emph{\textbf{Fixed Pose}} data. Results for significant findings ($p$-value $\leq 0.05$) are reported across the safety, comfort, and overall categories in Tables \ref{table:anova_results_safety}, \ref{table:anova_results_comfort}, \ref{table:anova_results_overall}, respectively. Phased reactivity has a statistically significant effect on all categories of
ratings, while transfer style has a statistically significant affect
on comfort ratings alone.

\begin{table}[!ht]
\centering
\setlength{\tabcolsep}{1.75pt}
\begin{tabular}{c | p{0.1\linewidth}}
\textbf{Variable} & $p$-value \\ 
\hline
Reactivity & \textbf{0.010802} \\
Transfer-Style & 0.949338\\
\end{tabular}
\makeatletter
\def\@captype{table}
\makeatother
\caption{\textbf{Overall: One-Way ANOVA Findings}}
\label{table:anova_results_overall}
\end{table}

\begin{table}[!ht]
\centering
\setlength{\tabcolsep}{1.75pt}
\begin{tabular}{c | p{0.1\linewidth}}
\textbf{Variable} & $p$-value \\ 
\hline
Reactivity & \textbf{0.000011} \\
Transfer-Style & \textbf{0.019153}\\
\end{tabular}
\makeatletter
\def\@captype{table}
\makeatother
\caption{\textbf{Comfort: One-Way ANOVA Findings}}
\label{table:anova_results_comfort}
\end{table}

\begin{table}[!ht]
\centering
\setlength{\tabcolsep}{1.75pt}
\begin{tabular}{c | p{0.1\linewidth}}
\textbf{Variable} & $p$-value \\ 
\hline
Reactivity & \textbf{0.000001} \\
Transfer-Style & 0.970068\\
\end{tabular}
\makeatletter
\def\@captype{table}
\makeatother
\caption{\textbf{Safety: One-Way ANOVA Findings}}
\label{table:anova_results_safety}
\end{table}

In Tables~\ref{table:fixed_ranking_values}--\ref{table:human_ranking_values}, we report the values of the 25th, 50th, and 75th percentile of user rankings on safety, comfort and overall bite quality per bite transfer method for each food axis category as shown in Figure~\ref{fig:grouped_rankings}. As in Figure~\ref{fig:grouped_rankings}, the 8 food items tested are grouped into three categories for each food variability axis (shape, size, and deformability). These groupings are reported in Table~\ref{table:food_groups}, which each food item appearing exactly once within each food variability axis classification.

\begin{table}[!ht]
\centering
\vspace{0.2cm}
\setlength{\tabcolsep}{1.75pt}
\begin{tabular}{c c | p{0.07\linewidth} p{0.07\linewidth} p{0.07\linewidth} | p{0.07\linewidth} p{0.07\linewidth} p{0.07\linewidth} | p{0.07\linewidth} p{0.07\linewidth} p{0.07\linewidth}}
\multirow{2}{*}{\textbf{Axis}} & \multirow{2}{*}{\textbf{Category}} & \multicolumn{3}{c|}{\textbf{Safety}} & \multicolumn{3}{c|}{\textbf{Comfort}} & \multicolumn{3}{c}{\textbf{Overall}} \\ 
 & & \emph{25th} & \emph{50th} & \emph{75th} & \emph{25th} & \emph{50th} & \emph{75th} & \emph{25th} & \emph{50th} & \emph{75th} \\ 
\hline
 & Round & 2 & 4 & 4 & 2 & 3 & 4 & 2 & 4 & 4 \\
Shape & Cube/Cyl & 2 & 4 & 4 & 2 & 2.5 & 4 & 2.75 & 3.5 & 4 \\
 & Irregular & 2.75 & 3.5 & 4.25 & 1.75 & 2.5 & 3.5 & 2.75 & 3 & 5 \\
\hline
 & Small & 2 & 4 & 4 & 2 & 3 & 4 & 2 & 4 & 4 \\
Size & Medium & 2 & 4 & 4 & 2 & 2 & 4 & 3 & 3 & 4 \\
 & Large & 2.5 & 4 & 4 & 2 & 3 & 4 & 2.5 & 3 & 4.5 \\
\hline
 & Fragile & 1 & 3 & 4 & 1 & 2 & 3 & 1 & 3 & 4 \\
Deform & Robust & 2 & 4 & 4 & 2 & 3 & 4 & 2 & 4 & 5 \\
 & Rigid & 3 & 4 & 4 & 2 & 4 & 4.5 & 3.5 & 4 & 4 \\

\end{tabular}
\makeatletter
\def\@captype{table}
\makeatother
\caption{\textbf{Fixed Pose User Ranking Values.}
\normalfont{
}}
\label{table:fixed_ranking_values}
\end{table}

\begin{table}[!ht]
\centering
\vspace{0.2cm}
\setlength{\tabcolsep}{1.75pt}
\begin{tabular}{c c | p{0.07\linewidth} p{0.07\linewidth} p{0.07\linewidth} | p{0.07\linewidth} p{0.07\linewidth} p{0.07\linewidth} | p{0.07\linewidth} p{0.07\linewidth} p{0.07\linewidth}}
\multirow{2}{*}{\textbf{Axis}} & \multirow{2}{*}{\textbf{Category}} & \multicolumn{3}{c|}{\textbf{Safety}} & \multicolumn{3}{c|}{\textbf{Comfort}} & \multicolumn{3}{c}{\textbf{Overall}} \\ 
 & & \emph{25th} & \emph{50th} & \emph{75th} & \emph{25th} & \emph{50th} & \emph{75th} & \emph{25th} & \emph{50th} & \emph{75th} \\ 
\hline
 & Round & 3 & 5 & 5 & 3 & 5 & 5 & 3 & 5 & 5 \\
Shape & Cube/Cyl & 3.75 & 5 & 5 & 4 & 5 & 5 & 3.75 & 5 & 5 \\
 & Irregular & 3.75 & 5 & 5 & 4.75 & 5 & 5 & 3.75 & 5 & 5 \\
\hline
 & Small & 3 & 5 & 5 & 3 & 5 & 5 & 3 & 5 & 5 \\
Size & Medium & 5 & 5 & 5 & 4 & 5 & 5 & 5 & 5 & 5 \\
 & Large & 2.5 & 5 & 5 & 3.5 & 5 & 5 & 2.5 & 5 & 5  \\
\hline
 & Fragile & 5 & 5 & 5 & 5 & 5 & 5 & 5 & 5 & 5 \\
Deform & Robust & 3 & 5 & 5 & 4 & 5 & 5 & 3 & 5 & 5  \\
 & Rigid & 2.5 & 5 & 5 & 3 & 4 & 5 & 2.5 & 5 & 5  \\

\end{tabular}
\makeatletter
\def\@captype{table}
\makeatother
\caption{\textbf{More Reactive User Ranking Values.}
\normalfont{
}}
\label{table:more_react_ranking_values}
\end{table}

\begin{table}[!ht]
\centering
\vspace{0.2cm}
\setlength{\tabcolsep}{1.75pt}
\begin{tabular}{c c | p{0.07\linewidth} p{0.07\linewidth} p{0.07\linewidth} | p{0.07\linewidth} p{0.07\linewidth} p{0.07\linewidth} | p{0.07\linewidth} p{0.07\linewidth} p{0.07\linewidth}}
\multirow{2}{*}{\textbf{Axis}} & \multirow{2}{*}{\textbf{Category}} & \multicolumn{3}{c|}{\textbf{Safety}} & \multicolumn{3}{c|}{\textbf{Comfort}} & \multicolumn{3}{c}{\textbf{Overall}} \\ 
 & & \emph{25th} & \emph{50th} & \emph{75th} & \emph{25th} & \emph{50th} & \emph{75th} & \emph{25th} & \emph{50th} & \emph{75th} \\ 
\hline
 & Round & 2 & 3 & 4 & 2 & 3 & 4 & 3 & 3 & 4 \\
Shape & Cube/Cyl & 2 & 5 & 5 & 3 & 4 & 4 & 2 & 4 & 4 \\
 & Irregular & 2.75 & 3.5 & 4 & 2.75 & 4 & 4 & 2 & 4 & 4 \\
\hline
 & Small & 2 & 3 & 4 & 2 & 3 & 4 & 3 & 3 & 4 \\
Size & Medium & 2 & 4 & 4 & 3 & 4 & 4 & 2 & 4 & 4 \\
 & Large & 2.5 & 3 & 4 & 2.5 & 4 & 4 & 2 & 4 & 4 \\
\hline
 & Fragile & 3 & 4 & 4 & 4 & 4 & 4 & 4 & 4 & 4 \\
Deform & Robust & 2 & 3 & 4 & 2 & 4 & 4 & 2 & 3 & 4  \\
 & Rigid & 2 & 2 & 3 & 2.5 & 3 & 3.5 & 2 & 2 & 3.5 \\

\end{tabular}
\makeatletter
\def\@captype{table}
\makeatother
\caption{\textbf{Less Reactive User Ranking Values.}
\normalfont{
}}
\label{table:less_react_ranking_values}
\end{table}

\begin{table}[!ht]
\centering
\vspace{0.2cm}
\setlength{\tabcolsep}{1.75pt}
\begin{tabular}{c c | p{0.07\linewidth} p{0.07\linewidth} p{0.07\linewidth} | p{0.07\linewidth} p{0.07\linewidth} p{0.07\linewidth} | p{0.07\linewidth} p{0.07\linewidth} p{0.07\linewidth}}
\multirow{2}{*}{\textbf{Axis}} & \multirow{2}{*}{\textbf{Category}} & \multicolumn{3}{c|}{\textbf{Safety}} & \multicolumn{3}{c|}{\textbf{Comfort}} & \multicolumn{3}{c}{\textbf{Overall}} \\ 
 & & \emph{25th} & \emph{50th} & \emph{75th} & \emph{25th} & \emph{50th} & \emph{75th} & \emph{25th} & \emph{50th} & \emph{75th} \\ 
\hline
 & Round & 2 & 3 & 3 & 2 & 3 & 3 & 2 & 3 & 3 \\
Shape & Cube/Cyl & 2 & 2.5 & 3.25 & 2 & 3 & 4 & 2 & 2 & 3 \\
 & Irregular & 2 & 2.5 & 3 & 3 & 3 & 3.25 & 2 & 3 & 3.25 \\
\hline
 & Small & 2 & 3 & 3 & 2 & 3 & 3 & 2 & 3 & 3 \\
Size & Medium & 2 & 2 & 3 & 2 & 3 & 3 & 2 & 2 & 3 \\
 & Large & 2 & 3 & 4 & 3 & 3 & 4.5 & 2.5 & 3 & 4 \\
\hline
 & Fragile & 2 & 2 & 3 & 2 & 3 & 3 & 2 & 2 & 2 \\
Deform & Robust & 2 & 3 & 3 & 2 & 3 & 3 & 2 & 3 & 3 \\
 & Rigid & 2 & 4 & 4.5 & 2 & 4 & 4.5 & 2 & 3 & 4.5 \\

\end{tabular}
\makeatletter
\def\@captype{table}
\makeatother
\caption{\textbf{OURS User Ranking Values.}
\normalfont{
}}
\label{table:ours_ranking_values}
\end{table}

\begin{table}[!ht]
\centering
\vspace{0.2cm}
\setlength{\tabcolsep}{1.75pt}
\begin{tabular}{c c | p{0.07\linewidth} p{0.07\linewidth} p{0.07\linewidth} | p{0.07\linewidth} p{0.07\linewidth} p{0.07\linewidth} | p{0.07\linewidth} p{0.07\linewidth} p{0.07\linewidth}}
\multirow{2}{*}{\textbf{Axis}} & \multirow{2}{*}{\textbf{Category}} & \multicolumn{3}{c|}{\textbf{Safety}} & \multicolumn{3}{c|}{\textbf{Comfort}} & \multicolumn{3}{c}{\textbf{Overall}} \\ 
 & & \emph{25th} & \emph{50th} & \emph{75th} & \emph{25th} & \emph{50th} & \emph{75th} & \emph{25th} & \emph{50th} & \emph{75th} \\ 
\hline
 & Round & 1 & 1 & 1 & 1 & 1 & 1 & 1 & 1 & 1 \\
Shape & Cube/Cyl & 1 & 2 & 3 & 1 & 2 & 2 & 1 & 2 & 3 \\
 & Irregular & 1 & 1 & 1.25 & 1 & 1 & 1.25 & 1 & 1 & 2 \\
\hline
 & Small & 1 & 1 & 1 & 1 & 1 & 1 & 1 & 1 & 1 \\
Size & Medium & 1 & 1 & 2 & 1 & 2 & 2 & 1 & 2 & 3 \\
 & Large & 1 & 2 & 3 & 1 & 1 & 2 & 1 & 1 & 2 \\
\hline
 & Fragile & 1 & 2 & 3 & 1 & 2 & 2 & 1 & 2 & 3 \\
Deform & Robust & 1 & 1 & 1 & 1 & 1 & 1 & 1 & 1 & 2 \\
 & Rigid & 1 & 2 & 3 & 1 & 2 & 2 & 1 & 2 & 2.5 \\

\end{tabular}
\makeatletter
\def\@captype{table}
\makeatother
\caption{\textbf{Human Oracle User Ranking Values.}
\normalfont{
}}
\label{table:human_ranking_values}
\end{table}

\begin{table}[!ht]
\centering
\vspace{0.2cm}
\begin{tabular}{c | c c c }
\textbf{Food} & \textbf{Shape} & \textbf{Size} & \textbf{Deformability} \\ 
\hline
Broccoli & Irregular & Large & Robust \\
Blueberry & Round & Small & Robust \\
Cherry Tomato & Round & Small & Robust \\
Strawberry & Irregular & Large & Robust \\
Carrot & Cylinder & Large & Rigid \\
Pineapple & Cube & Medium & Rigid \\
Tofu & Cube & Medium & Fragile \\
Cheesecake & Cube & Medium & Fragile \\
\end{tabular}
\makeatletter
\def\@captype{table}
\makeatother
\caption{\textbf{Food Properties:}
\normalfont{We group the 8 food items we consider along three axes of variability: shape, size, and deformability. We consider three classes for each axis, as shown in Figure~\ref{fig:grouped_rankings}. For shape, we define an ``irregularly shaped food'' to be one with less than three axes of symmetry.  
}}
\label{table:food_groups}
\end{table}

\begin{figure*}[!htbp] 
\centering
\includegraphics[width=\linewidth]{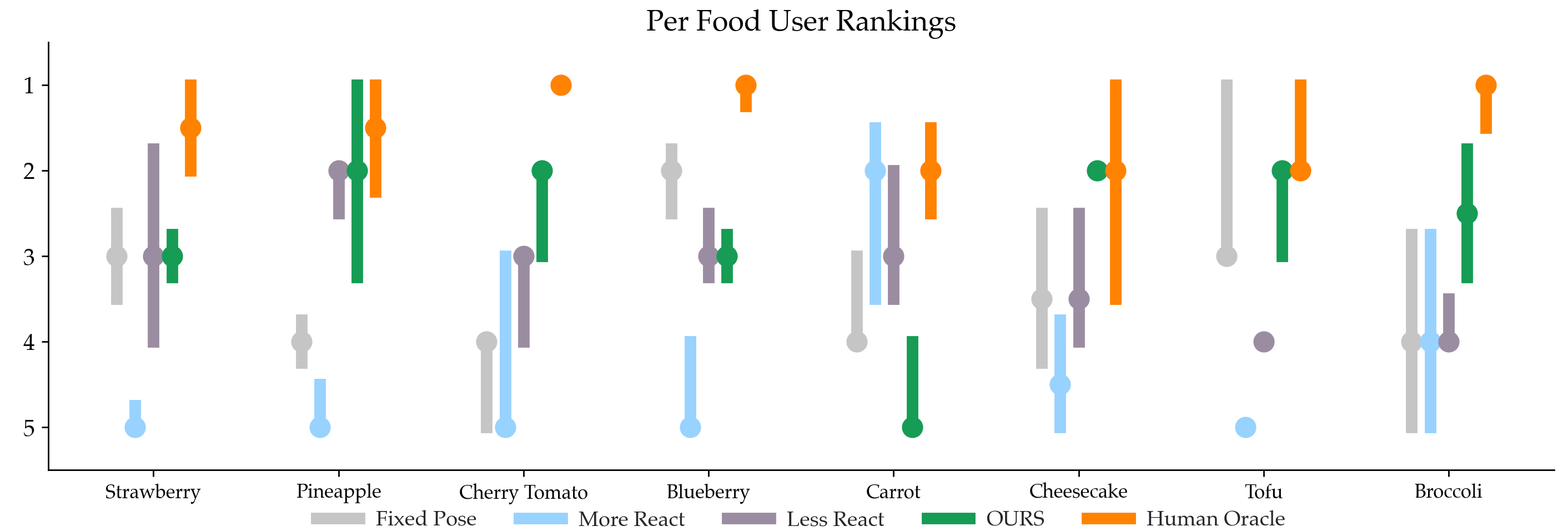}
\caption{\textbf{Per Food User Rankings.} We report the median and interquartile range of user rankings of the overall bite quality for our bite transfer method and four baselines for each of the 8 food groups considered. 1 indicates best performance and 5 indicates worst. We observe that our method performs as well or better than all robotic feeding baselines acorss all foods except carrots. We highlight that our method approaches human feeding oracle performance in the pineapple and cheesecake food classes, supporting the effectiveness of our robotic feeding method.}
\label{fig:per_food_rankings}
\end{figure*}

Figure~\ref{fig:per_food_rankings} reports the user rankings for bite transfer methods for each of the 8 foods considered. These results show that our method either outperforms or performs comparably to the More React and Less React baselines in all foods except carrot. 
Because the carrot is slightly larger than the size of some users mouths, some users preferred more autonomy compared to other foods to adjust the positioning of the carrot and fork in their mouth. 
As a result, some users preferred the More Reactive baseline over our method as the user could adjust the fork and carrot positioning more easily.
Some users also preferred the Less React baseline over our method because the rigidity of the carrot complicates biting off a fork with our method which has a higher reactivity in comparison.
As expected, the Human Oracle baseline achieves the highest performance of all the methods across all food items. The Fixed Pose baseline has widely varied performance across the food classes, suggesting some foods, such as tofu cubes and blueberries, are easier to bite off a fixed fork pose, while the other foods are easier to eat when delivered into the user mouth.

\begin{figure}[!htbp]
\centering
\includegraphics[width=\linewidth]{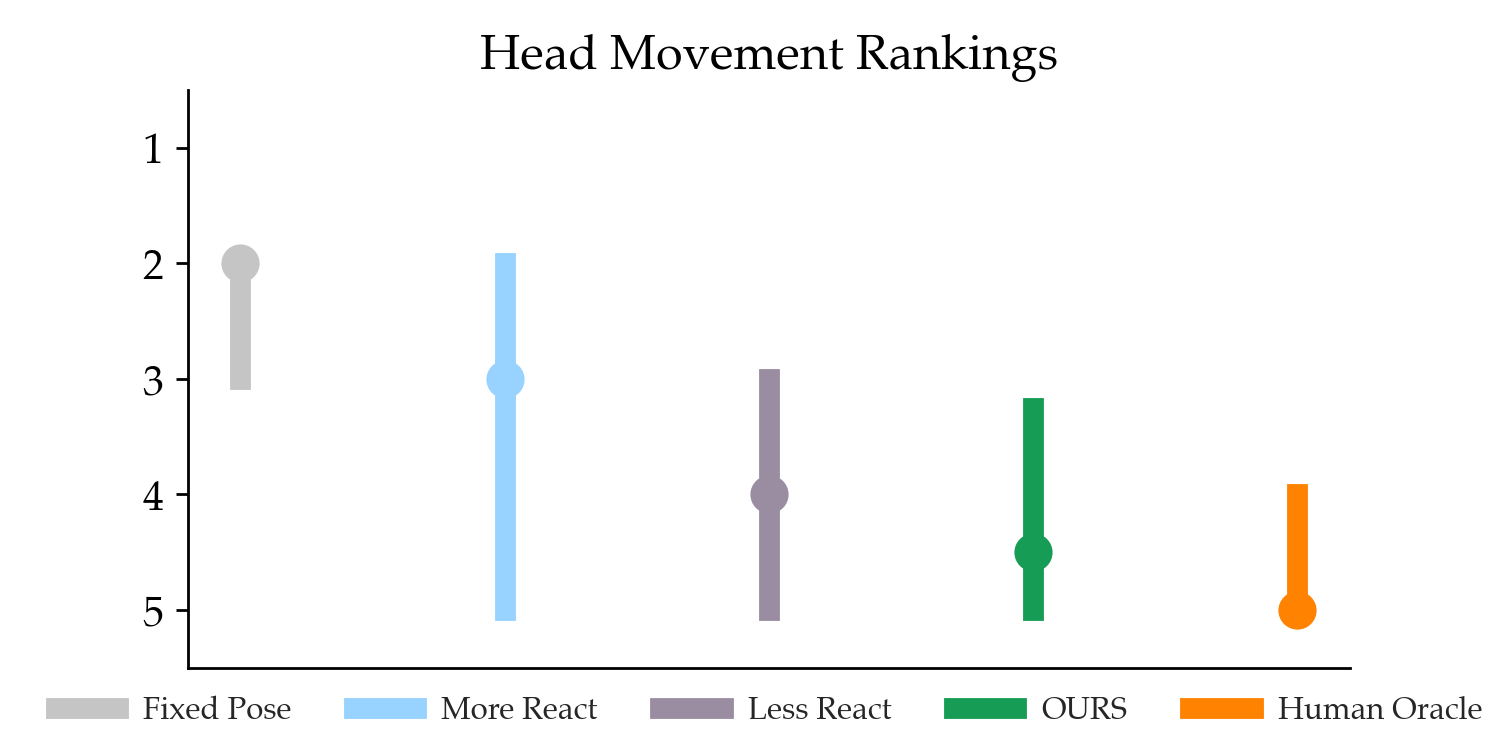}
\caption{\textbf{Head Movement User Rankings.} We report the user rankings for the ease with which they could take a bite with each bite transfer method without moving their head. 1 signifies the user had to move more to take a bite, while 5 signifies less head movement. We find that the Fixed Pose baseline required the most head movement due to the nature of the method holding the food item outside the mouth for the user to take a bite. }
\label{fig:head_move_rankings}
\end{figure}

We report the user rankings of head movement required for each bite transfer method in Figure~\ref{fig:head_move_rankings}. We highlight the Fixed Pose method required the most head movement from the user, due to the nature of the fork approaching the mouth position, but stopping outside instead of attempting an in-mouth transfer. The three in-mouth robotic bite transfer methods (More React, Less React, and Ours) are all ranked similarly for required head movement, which is expected as these methods use the same transfer trajectory to achieve in-mouth delivery.
Additionally, we note that our bite transfer method approaches the Human Oracle performance in terms of amount of head movement required.

\end{document}